\documentclass[letterpaper, 10 pt, conference]{ieeeconf}
\usepackage{booktabs}
\usepackage{balance}
\usepackage{cite}
\usepackage{amsmath,amssymb,amsfonts}
\usepackage{algorithm}
\usepackage{algpseudocode}
\usepackage{graphicx}
\usepackage{textcomp}
\usepackage{xcolor}
\usepackage{verbatim}
\usepackage{graphicx}
\usepackage{multirow}
\usepackage{mathrsfs}

\IEEEoverridecommandlockouts
\begin{document}

\title{VLM-RRT: Vision Language Model Guided RRT Search \\ for Autonomous UAV Navigation}
\author{Jianlin~Ye,~Savvas~Papaioannou and Panayiotis~Kolios
\thanks{The authors are with the KIOS Research and Innovation Centre of Excellence (KIOS CoE), and the Department of Computer Science, University of Cyprus, Nicosia, 1678, Cyprus. {\tt\small \{ye.jianlin,  papaioannou.savvas, pkolios\}@ucy.ac.cy}}
}
\maketitle

\begin{abstract}
Path planning is a fundamental capability of autonomous Unmanned Aerial Vehicles (UAVs), enabling them to efficiently navigate toward a target region or explore complex environments while avoiding obstacles. Traditional path-planning methods, such as Rapidly-exploring Random Trees (RRT), have proven effective but often encounter significant challenges. These include high search space complexity, suboptimal path quality, and slow convergence, issues that are particularly problematic in high-stakes applications like disaster response, where rapid and efficient planning is critical. To address these limitations and enhance path-planning efficiency, we propose Vision Language Model RRT (VLM-RRT), a hybrid approach that integrates the pattern recognition capabilities of Vision Language Models (VLMs) with the path-planning strengths of RRT. By leveraging VLMs to provide initial directional guidance based on environmental snapshots, our method biases sampling toward regions more likely to contain feasible paths, significantly improving sampling efficiency and path quality. Extensive quantitative and qualitative experiments with various state-of-the-art VLMs demonstrate the effectiveness of this proposed approach.
\end{abstract}

\section{Introduction} 
\label{sec:Introduction}

As Unmanned Aerial Vehicles (UAVs) operate in increasingly dynamic and complex environments, the demand for reliable navigation \cite{vitale2022autonomous}, including efficient and adaptive path-planning strategies \cite{vitale2024probabilistically}, has grown significantly. Path planning, a critical component of autonomous UAV navigation, determines the optimal path from a starting point to a target region while avoiding obstacles, often optimizing specific mission objectives \cite{lavalle2006planning,papaioannou2023distributed,10925885,papaioannou2025rolling}. This process is central to applications such as emergency response \cite{wan2022accurate, papaioannou20213d, papaioannou2021towards,papaioannou2020coordinated,papaioannou2024synergising}, surveillance \cite{papaioannou2022distributed,wu2020cooperative, papaioannou2019probabilistic, papaioannou2020cooperative,papaioannou2023joint}, and automated inspection \cite{jing2020multi, sun2022automating,papaioannou2022uav,PAPAIOANNOU2024532,papaioannou2023cooperative,papaioannou2023unscented}.

Existing sampling-based path-planning algorithms, such as Rapidly-exploring Random Trees (RRT) \cite{lavalle1998rapidly, kuffner2000rrt}, offer significant advantages, including their ability to handle high-dimensional spaces and their probabilistic completeness, meaning they will eventually find a solution if one exists. However, these methods require careful fine-tuning and often fail to converge to optimal solutions, particularly in complex or cluttered environments. This limitation is critical, as it reduces their reliability, making them less suitable for high-stakes applications like search-and-rescue and disaster response missions, where rapid and dependable solutions are paramount. Recent hybrid approaches combining sampling-based methods with machine learning techniques have shown promise in addressing these shortcomings, offering improved computational efficiency and path quality \cite{arslan2015machine, wang2021deep, kuo2018deep}.

In this direction, this work integrates multimodal large language models (LLMs) with RRT-based path planning to tackle these challenges. The proposed framework, Vision Language Model RRT (VLM-RRT), combines the strengths of sampling-based path planning through RRT with the pattern-matching capabilities and emergent reasoning of LLMs, thereby enhancing autonomous UAV navigation by enabling efficient and robust path planning. Specifically, the proposed approach incorporates a VLM module into the path-planning process to analyze environmental snapshots and dynamically guide the planner to prioritize and sample from specific regions. This reduces redundant exploration and accelerates convergence by biasing the sampling process toward regions more likely to contain feasible and efficient paths, thus improving convergence rates and path quality. The contributions of this work can be summarized as follows:

\begin{itemize}
\item We propose VLM-RRT, a novel framework that augments traditional RRT-based path planning with the advanced reasoning capabilities of Vision Language Models (VLMs). By leveraging a VLM module to analyze environmental snapshots, VLM-RRT enhances navigation decisions by dynamically guiding the sampling process, effectively biasing it toward regions with a higher likelihood of containing optimal paths.
\item Extensive qualitative and quantitative experimental evaluations using OpenAI's GPT-4o and Meta's Llama 3.2 multimodal LLMs demonstrate the effectiveness of the proposed approach in terms of sampling efficiency and path quality compared to the standalone RRT approach.
\end{itemize}

The remainder of this paper is organized as follows. Section \ref{sec:Related_Work} provides background and discusses related work, Section \ref{sec:Problem} formulates the problem, and Section \ref{sec:Approach} details the proposed approach. Finally, Section \ref{sec:Evaluation} evaluates the proposed approach, and Section \ref{sec:Conclusion} concludes the paper.

\section{Related Work}
\label{sec:Related_Work}

The Rapidly-exploring Random Tree (RRT) algorithm \cite{lavalle1998rapidly} is a sampling-based method for path planning, designed to efficiently explore high-dimensional configuration spaces. It incrementally builds a tree by randomly sampling points in the space and connecting them to the nearest point in the existing tree, ensuring rapid exploration. RRT is particularly effective for finding feasible paths in complex, obstacle-filled environments, making it a widely used approach in robotics and autonomous navigation.

Significant algorithmic refinements have enhanced RRT's capabilities over the years. For instance, the RRT* algorithm \cite{karaman2011sampling} introduced asymptotic optimality through node-rewiring mechanisms by minimizing a predefined cost function (e.g., path length), while Informed RRT* \cite{gammell2014informed} developed advanced sampling strategies to guide exploration near optimal solution regions. On the other hand, RRT-Connect \cite{kuffner2000rrt} grows two trees bidirectionally from both the start and goal configurations. This approach, combined with a greedy heuristic, allows faster exploration of the configuration space and quicker convergence to a solution compared to the standard RRT.

To tackle issues related to random sampling and low path efficiency in RRT, an improved approach incorporating the Artificial Potential Field (APF) method was recently proposed in \cite{huang2021path}. This method introduces a probability value during the expansion step of the random tree in the basic RRT algorithm, enhancing convergence speed toward the target node. Similarly, the authors in \cite{app14010025} proposed an RRT variant that utilizes adjustable probability and sampling area strategies to quickly find a feasible path. This planner is then combined with an optimizer that uses the Dijkstra algorithm to prune and improve the initial path. More recently, the work in \cite{liu2024rrt} combined RRT with model predictive control (MPC) utilizing control barrier functions (CBF) to enforce safety-critical constraints. Additionally, recent advancements in learning-based RRT methods have shown promising results in improving path-planning efficiency. The Neural Informed RRT* approach \cite{huang2024neural} utilizes a neural network to learn the topology of the free space and infer states close to the optimal path, thereby guiding the search toward more promising regions, whereas Neural RRT* \cite{wang2020neural} employs convolutional neural networks to predict a probabilistic heatmap of states for guiding exploration.

Finally, the emergence of Large Language Models (LLMs) \cite{touvron2023llama, brown2020language} has revolutionized artificial intelligence, enabling advanced reasoning and knowledge-driven applications in autonomous navigation. Recent research has explored the potential of LLMs to enhance navigation tasks by combining their reasoning capabilities with domain-specific methods. For instance, the work in \cite{zhou2024navgpt} demonstrated that LLMs can perform high-level planning tasks for navigation, including identifying landmarks from observed scenes, tracking navigation progress, and correcting course. More closely related to our work is the approach in \cite{meng2024llm}, where the authors integrated LLMs with the A* path-planning algorithm \cite{Hart1968}, achieving enhanced pathfinding efficiency in terms of time and space complexity.

In summary, RRT-based path-planning approaches are extensively utilized for their ability to explore state spaces efficiently and effectively. However, they often require meticulous parameter tuning and tend to exhibit slow convergence. While these methods can find optimal solutions, they frequently incur significant computational overhead, with high memory and time demands, particularly when searching for the best path. This limitation is especially critical in applications such as autonomous vehicles, where rapid identification of efficient paths is essential due to constraints like limited power or fuel resources. Motivated by these challenges, this work proposes a novel path-planning framework that integrates the reasoning capabilities of multimodal large language models with RRT-based path search to enhance the efficiency of path generation.

\section{Preliminaries}
\label{sec:Problem}
This study addresses a UAV path-planning challenge inspired by real-world wildfire disaster response scenarios. The objective is to autonomously navigate a UAV through fire-affected forested regions to locate survivors while ensuring safe traversal by avoiding hazardous fire fronts. The UAV must reach a predefined goal region while dynamically adapting its trajectory to evolving environmental conditions.
To support this task, we assume the presence of a disaster early-warning system (EWS) equipped with multimodal sensing capabilities, including satellite imagery and meteorological data. This system provides real-time updates on the locations of both fire fronts and survivors, enabling the UAV to maintain up-to-date situational awareness. Leveraging this information, the UAV must compute an optimal trajectory that balances mission success with environmental constraints, ensuring efficient and safe navigation through the disaster zone.

\subsection{UAV Dynamical Model} 
\label{ssec:agent_dynamics}
Without loss of generality, we assume that the dynamical behavior $\mathscr{B}$ of a UAV agent can be described by a linear time-invariant (LTI) system \cite{how2015linear} of the following form:
\begin{equation} \label{eq:LTI}
\mathscr{B}(A,B,C,D) := 
\begin{cases}
x(t + 1) = Ax(t) + Bu(t) \\
y(t) = Cx(t) + Du(t),
\end{cases}
\end{equation}
where $\mathscr{B}(A,B,C,D)$ is the input/output/state representation of the system, with \( A \in \mathbb{R}^{n \times n} \), \( B \in \mathbb{R}^{n \times m} \), \( C \in \mathbb{R}^{p \times n} \), and \( D \in \mathbb{R}^{p \times m} \)  known. The state, control input, and output of the system at time-step \(t \in \mathbb{N}\) are given respectively by \( x(t) \in \mathcal{X} \subset \mathbb{R}^n \), \( u(t) \in \mathcal{U} \subset \mathbb{R}^m \),  and \( y(t) \in \mathcal{Y} \subset \mathbb{R}^p \). 
For conciseness, we assume that \( x(t) \in \mathbb{R}^6 \) represents the 3D position and linear velocity of the UAV in Cartesian coordinates. The control input force is denoted by \( u(t) \in \mathbb{R}^3 \), and the UAV position by \( y(t) \in \mathbb{R}^3 \).

\subsection{Environment Model} 
\label{ssec:env_model}
The UAV operates within a bounded three-dimensional environment $\mathcal{E} \subset \mathbb{R}^3$, which consists of (a) a designated goal region $\mathcal{G}$ that the UAV must reach during its mission and (b) a set of fire fronts $\mathcal{O}$ that must be avoided. The mission begins at the UAV's home depot $\mathcal{S}$, from which it departs at the start of the search and concludes upon reaching $\mathcal{G}$. All relevant environmental information, including the locations of $\mathcal{S}$, $\mathcal{G}$, and fire fronts $o \in \mathcal{O}$, is assumed to be fully known and provided to the UAV at the outset of the mission. In this formulation, $\mathcal{S}$, $\mathcal{G}$, and the set of fire fronts $\mathcal{O}$ are represented as rectangular cuboids of varying dimensions.

\subsection{Problem Formulation} 
\label{ssec:formulation}

The autonomous UAV navigation problem, as previously discussed, can be formulated as a finite-horizon optimal control problem, as presented in Eq.~\eqref{eq:MPC}. The objective is to determine the optimal UAV control inputs \( u(t) \), for \( t \in \{0, \dots, T-1\} \), over a suitably chosen planning horizon of \( T \) time steps. These control inputs guide the UAV toward the goal region \( \mathcal{G} \) by tracking a reference path \( \mathcal{P} \) while adhering to the system's dynamics and constraints, as shown.

\vspace{-2mm}
\begin{equation}\label{eq:MPC}
\begin{aligned}
    \min_{u, y} \quad & \sum_{t=0}^{T-1} \left( \| y(t) - \mathcal{P}(t) \|_Q^2 + \| u(t) \|_R^2 \right) \\
    \text{s.t.} \quad & x(t+1) = A x(t) + B u(t), \quad \forall t \in \{0, \dots, T-1\}, \\
    & y(t) = C x(t) + D u(t), \quad \forall t \in \{0, \dots, T-1\}, \\
    & x(0) = x_{init}, \\
    & x(t) \in \mathcal{X}, \quad \forall t \in \{0, \dots, T-1\}, \\
    & u(t) \in \mathcal{U}, \quad \forall t \in \{0, \dots, T-1\}, \\
    & y(t) \in \mathcal{Y}, \quad \forall k \in \{0, \dots, N-1\}.
\end{aligned}
\end{equation}

\noindent In the formulation above, \( x_{\text{init}} \) represents the agent's initial state, positioning the agent within its home depot \( \mathcal{S} \). The objective is to track the reference path \( \mathcal{P}(t) \), \( t \in \{0, \ldots, T-1\} \) (which describes a feasible path to the goal region), over the time horizon, subject to the UAV's dynamical constraints and operational behavior. The norm \( \|u(t)\|^2_R \) denotes the quadratic form \( u(t)^\top R u(t) \) (likewise for \( \|\cdot\|^2_Q \)), where \( R \in \mathbb{R}^{m \times m} \) is the control cost matrix and \( Q \in \mathbb{R}^{p \times p} \) is the output cost matrix. In essence, the objective function minimizes the weighted sum of the tracking error (i.e., deviation from the reference path) and control effort, with the weighting scheme captured in the matrices \( Q \) and \( R \), respectively. The reference path \( \mathcal{P} \) provides a collision-free path from \( \mathcal{S} \) to the goal region \( \mathcal{G} \). Once \( \mathcal{P} \) is known, the solution to the optimization problem in Eq.~\eqref{eq:MPC} can be obtained using numerical optimization, such as quadratic programming \cite{nocedal2006quadratic}. In the following section, we present how the proposed VLM-RRT approach generates the reference path \( \mathcal{P} \) by integrating a Large Language Model with the Rapidly-exploring Random Tree algorithm. To simplify the analysis without loss of generality, we assume that the UAV operates at a fixed altitude, thereby restricting our formulation to a planar 2D setting. However, the proposed approach can be readily extended to three-dimensional navigation.

\begin{algorithm}[ht]
\caption{Traditional RRT Algorithm}
\label{alg:traditional_rrt}
\begin{algorithmic}[1]
\Require $y(0), ~\mathcal{G}_o,~\mathcal{O}$, $\delta$
\State $V \gets \{y(0)\}$, $E \gets \emptyset$, $r \gets \emptyset$, $i \gets 0$
\While{$i < N$}
    \State $\nu_{\text{rand}} \gets \text{SampleState()}$
    \State $\nu_{\text{nearest}} \gets \text{NearestNeighbor}(V, \nu_{\text{rand}})$
    \State $\nu_{\text{new}} \gets \text{Steer}(\nu_{\text{nearest}}, \nu_{\text{rand}}, \delta)$
    \State $i \gets i + 1$
	\If{$\text{PathFree}(\nu_{\text{new}},\nu_{\text{nearest}}, \mathcal{O})$}
        \State $V \gets V \cup \{\nu_{\text{new}}\}$, $E \gets E \cup \{(\nu_{\text{nearest}}, \nu_{\text{new}})\}$
    \EndIf
    \If{$||\nu_{\text{new}} -  \mathcal{G}_o||_2 \leq \epsilon$}
        \State $\mathcal{P} \gets \text{RetrievePlan}(V, E, \nu_{\text{new}})$
        \State \Return $\mathcal{P}$
    \EndIf
\EndWhile
\end{algorithmic}
\end{algorithm}


\section{Proposed Approach}
\label{sec:Approach}

Traditionally, we can obtain the reference path \( \mathcal{P} \) using the RRT sampling-based path-planning algorithm, depicted in Alg.~\ref{alg:traditional_rrt}. As demonstrated, the algorithm receives as input the initial position of the agent, denoted as \( y(0) \), at time \( t = 0 \), along with the goal region \( \mathcal{G}_o \in \mathbb{R}^2 \) (e.g., the centroid of the target region \( \mathcal{G} \)) and the set of fire fronts \( \mathcal{O} \). The algorithm proceeds by incrementally constructing a tree \( V \), originating from the agent's initial state and expanding toward the goal region \( \mathcal{G}_o \). During each iteration, a point \( \nu_{\text{rand}} \) is randomly sampled from the space, and the tree is extended from its closest existing vertex \( \nu_{\text{nearest}} \) toward \( \nu_{\text{rand}} \), resulting in a new vertex \( \nu_{\text{new}} \), provided that the path does not intersect any fire fronts, ensuring navigational safety. The extension follows a predefined step size \( \delta \), systematically guiding the tree's exploration of the space. This iterative process continues until either the tree successfully reaches the goal region (i.e., \( \|\nu_{\text{new}} - \mathcal{G}_o\|_2 \leq \epsilon \), where \( \epsilon > 0 \)), in which case the path is retrieved via backtracking, or the predefined iteration limit \( N \) is reached, resulting in the algorithm failing to converge.

The RRT algorithm, while widely effective across various motion planning tasks, encounters several inherent challenges that can impact its performance. One significant limitation is low sampling efficiency, as traditional RRT often produces a high proportion of invalid or redundant samples. This inefficiency increases computational overhead and hampers the algorithm's ability to explore the search space effectively. Additionally, the resulting paths may include unnecessary detours or redundant nodes, which can lead to suboptimal path quality and increased traversal cost.

To address these limitations, VLM-RRT utilizes LLMs as general-purpose pattern-matching machines and integrates their reasoning capabilities into the RRT search to guide the sampling process toward regions most likely to contain the optimal solution.

\begin{algorithm}
\caption{VLM-RRT Algorithm}
\label{alg:vlm_rrt}
\begin{algorithmic}[1]
\Require $y(0), ~\mathcal{G}o,~\mathcal{O}$, $\delta, \gamma$
\State $V \gets {y(0)}$, $E \gets \emptyset$, $r \gets \emptyset$, $i \gets 0$
\While{$i < N$}
\State $E_{\text{current}} \gets \text{GetEnvironmentState}()$
\State $\alpha \gets \mathcal{U}(0,1)$ 
\If{$\alpha \leq \gamma$}
\State	$\hat{\nu} \gets \text{PickLeafNode}(V,E_{\text{current}})$
\State $d \gets \text{GetVLMdirection}(\hat{\nu},E_{\text{current}})$
\State $\nu_{\text{rand}} \gets \text{SampleStateVLM}(\hat{\nu},d, r, \theta)$
\Else
\State $\nu_{\text{rand}} \gets \text{SampleState}()$
\EndIf
\State $\nu_{\text{nearest}} \gets \text{NearestNeighbor}(V, \nu_{\text{rand}})$
\State $\nu_{\text{new}} \gets \text{Steer}(\nu_{\text{nearest}}, \nu_{\text{rand}}, \delta)$
\State $i \gets i + 1$
\If{$\text{PathFree}(\nu_{\text{new}},\nu_{\text{nearest}}, \mathcal{O})$}
\State $V \gets V \cup {\nu_{\text{new}}}$, $E \gets E \cup {(\nu_{\text{nearest}}, \nu_{\text{new}})}$
\EndIf
\If{$||\nu_{\text{new}} -  \mathcal{G}_o||_2 \leq \epsilon$}
\State $\mathcal{P} \gets \text{RetrievePlan}(V, E, \nu_{\text{new}})$
\State \Return $\mathcal{P}$
\EndIf
\EndWhile
\end{algorithmic}
\end{algorithm}

\begin{figure*}
    \centering
    \includegraphics[width=1.0\textwidth]{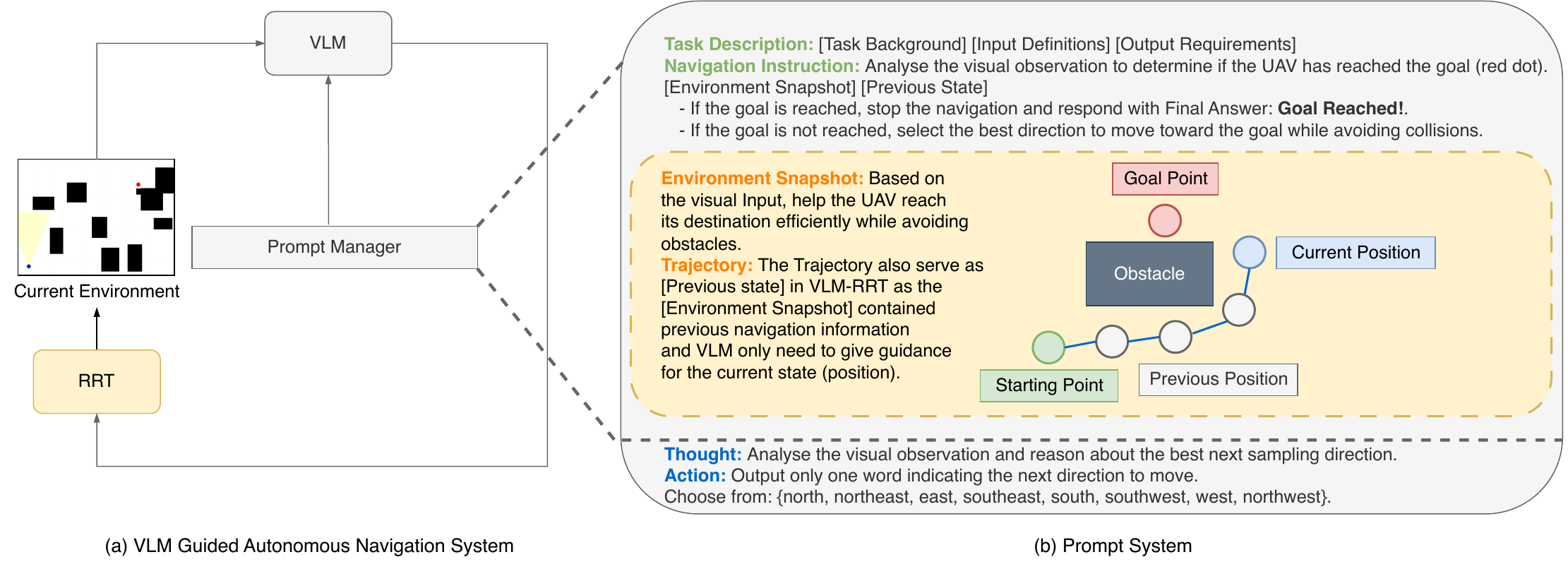}
    \caption{Our basic system consists of two types of prompts, task descriptions and basic inputs. We match a snapshot of the current environment with the task instructions, incorporating the current navigation state and history into the prompt to activate the agent's global dynamic exploration capability.}
    \label{fig:system_achitecture}
\end{figure*}

\subsection{VLM-RRT}
\label{ssec:vlm_rrt}

The VLM-RRT algorithm, shown in Alg.~2, integrates VLMs into the sampling process of RRT. In this framework, VLMs serve as general-purpose pattern-matching and reasoning machines that extract contextual information from the current environment, as illustrated in Fig.~\ref{fig:system_achitecture}. The VLM output aims to improve sampling efficiency and steer the tree exploration toward regions that are more likely to yield an optimal path. In our context, ``optimal'' refers to the collision-free route that minimizes the total travel distance from the starting position to the goal region, ensuring the most efficient navigation.

At the start of each planning iteration, the algorithm captures the current state of the environment as an image \( E_{\text{current}} \) using the function \texttt{GetEnvironmentState}. This image encodes the locations of fire fronts \( \mathcal{O} \), the goal region, and the state of exploration represented by the tree \( V \). In this representation, the goal region and the leaf nodes in \( V \) are distinguished using different colors.

Subsequently, with probability \( \gamma \), the algorithm decides whether to take a VLM-informed exploration step. In Line 4, the random variable \( \alpha \) is drawn from the uniform distribution in the range \( (0,1) \). With probability \( \gamma \), the proposed approach randomly selects a leaf node \( \hat{\nu} \) from \( V \), as shown in Line 6 of Alg.~2, and then employs the VLM to determine the direction \( d \) in which the agent should move to reach the goal region, given the selected node \( \hat{\nu} \). This is achieved through the function \texttt{GetVLMdirection}, as shown in Line 7, which leverages the VLM via prompt engineering to detect the goal region and reason about the direction \( d \) the agent should take by analyzing the environment image \( E_{\text{current}} \).

The algorithm then proceeds by randomly sampling a new point \( \nu_{\text{rand}} \) from a sector \( \mathcal{R} \), centered at \( \hat{\nu} \), with direction \( d \), radius \( r \), and angle \( \theta \), as described in Line 8. Otherwise, with probability \( 1 - \gamma \), the algorithm follows the standard RRT sampling strategy, selecting \( \nu_{\text{rand}} \) from anywhere in the environment using the function \texttt{SampleState}, as shown in Line 10. Subsequently, the algorithm operates similarly to the original RRT, where the tree is expanded from its nearest existing vertex, \( \nu_{\text{nearest}} \), in the direction of \( \nu_{\text{rand}} \), generating a new vertex \( \nu_{\text{new}} \). This expansion occurs only if the resulting path does not intersect any fire fronts, thereby ensuring safe navigation toward the goal region. This iterative process continues until either the tree successfully reaches the goal region or the algorithm reaches the predefined iteration limit.

\begin{figure*}
    \centering
    \includegraphics[width=1.0\textwidth]{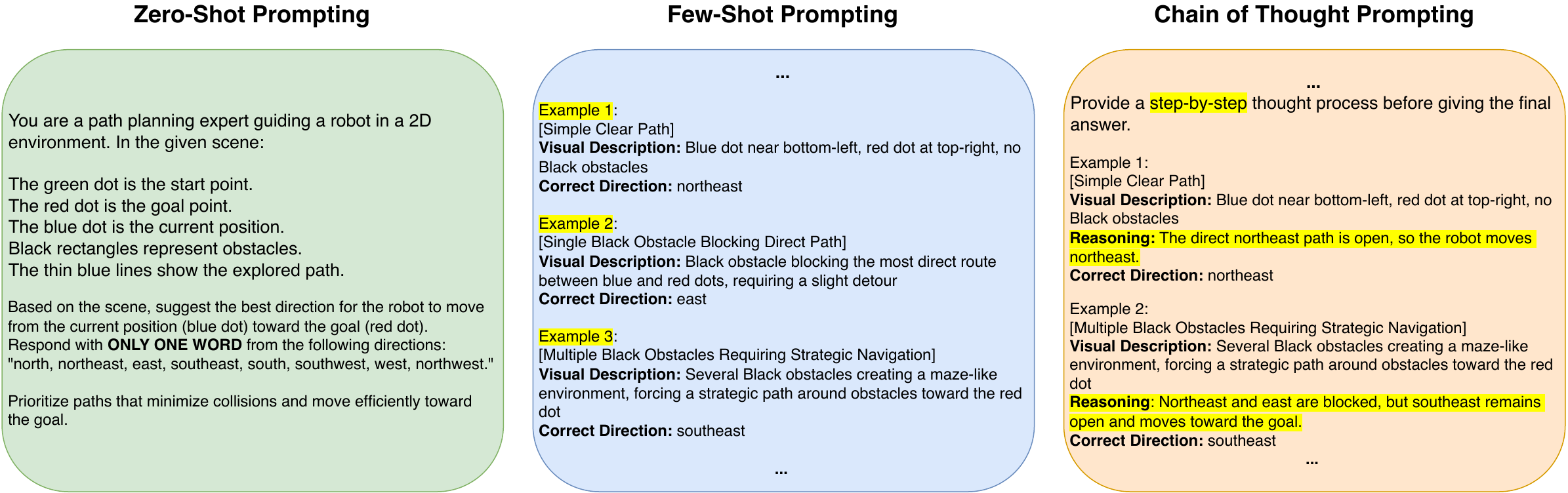}
    \caption{Comparison of different prompt engineering techniques for navigation decision-making.}
    \label{fig:prompt}
\end{figure*}

\subsection{Prompt Engineering}

The prompt engineering methodology leverages the VLM's pattern-matching capabilities. As shown in Fig.~\ref{fig:system_achitecture}, our system implements structured prompts that combine task descriptions with environmental snapshots, incorporating both current navigation states and historical data. The prompt structure defines specific input parameters, output constraints, and environmental context for navigation guidance.

As demonstrated in Fig.~\ref{fig:prompt}, we experimented with three prompting techniques. Zero-shot prompting enables direct decision-making using only task instructions and current state information. Few-shot prompting augments this by including predefined example scenarios, ranging from unobstructed paths to multi-obstacle configurations, which serve as reference cases for similar navigation contexts. We integrated Chain of Thought (CoT) prompting~\cite{wei2022chain} to enhance the system's reasoning capabilities. The CoT framework structures the decision-making process into explicit steps: obstacle identification, relative position analysis, and path feasibility evaluation. This structured approach enables the model to systematically process environmental constraints before determining movement directions.

\section{Evaluation}
\label{sec:Evaluation}

\begin{figure*}[!htb]
    \centering
    \includegraphics[width=1.0\textwidth]{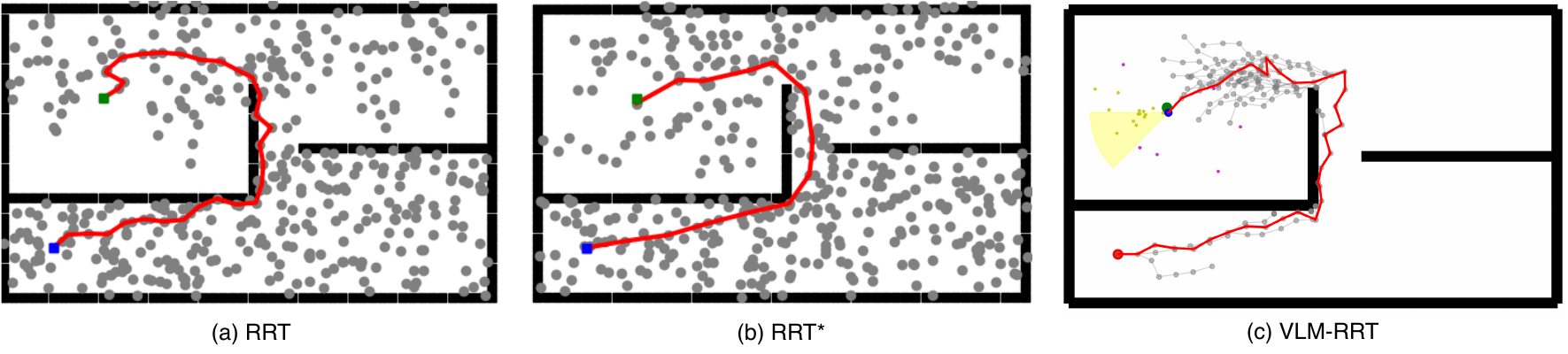}
    \caption{Illustrative example of the path-planning behavior obtained with: (a) RRT, (b) RRT* and (c) VLM-RRT.}
    \label{fig:Baseline_image}
\end{figure*}

\subsection{Simulation Setup} 
\label{ssec:sim_setup}

To evaluate our approach, we assume that an autonomous UAV agent evolves (assuming a fixed altitude) inside a bounded environment \( \mathcal{E} \subset \mathbb{R}^2 \) of dimensions \( 500 \, \text{m} \times 500 \, \text{m} \). The agent's planar motion is captured by a 4-dimensional state vector \( x(t) = [x_1, x_2, \dot{x}_1, \dot{x}_2]^\top \in \mathcal{X} \subset \mathbb{R}^4 \), comprising its position \( (x_1, x_2) \in \mathbb{R}^2 \) and velocity \( (\dot{x}_1, \dot{x}_2) \in \mathbb{R}^2 \) components within the 2D Cartesian coordinate system. The agent is controllable and capable of following specific directional and speed commands via the control input \( u(t) \in \mathcal{U} \subset \mathbb{R}^2 \), which corresponds to the applied control force. The matrices \( A \in \mathbb{R}^{4 \times 4} \) and \( B \in \mathbb{R}^{4 \times 2} \), shown in Eq.~\eqref{eq:LTI}, are given by:
\begin{equation*}
A = 
\begin{bmatrix}
    I_{2 \times 2} & \Delta T \cdot I_{2 \times 2} \\
    0_{2 \times 2} & (1 - \zeta) \cdot I_{2 \times 2}
\end{bmatrix}, \quad B = 
\begin{bmatrix}
    0_{2 \times 2} \\
    \frac{\Delta T}{m} \cdot I_{2 \times 2}
\end{bmatrix},
\end{equation*}
where \( \Delta T \) signifies the sampling interval, \( \zeta \) is the air resistance coefficient, and \( m \) represents the mass of the agent. Additionally, \( I_{2 \times 2} \) and \( 0_{2 \times 2} \) are the 2-by-2 identity and zero matrices, respectively. The output vector \( y(t) \in \mathcal{Y} \subset \mathbb{R}^2 \) consists of the UAV's position at time step \( t \); therefore, the matrices \( C \in \mathbb{R}^{2 \times 4} \) and \( D \in \mathbb{R}^{2 \times 2} \) are given by \( \begin{bmatrix} I_{2 \times 2} & 0_{2 \times 2} \end{bmatrix} \) and \( 0_{2 \times 2} \), respectively. The parameters \( \Delta T \), \( \zeta \), and \( m \) are set to \( 1 \, \text{s} \), \( 0.2 \), and \( 1.05 \, \text{kg} \), respectively. The control input is bounded in each dimension within the range \( [-10, 10] \, \text{N} \), and the UAV's maximum velocity is capped at \( v_{\text{max}} = 15 \, \text{m/s} \). The starting and goal regions \( \mathcal{S} \) and \( \mathcal{G} \), as well as the fire fronts to be avoided \( o \in \mathcal{O} \), are represented as rectangular regions with random dimensions, as shown in Fig.~\ref{fig:system_achitecture}(a).

Unless otherwise indicated, the VLM-RRT step size \( \delta \) is set to \( \Delta T \cdot v_{\text{max}} \) and \( \epsilon = 1 \, \text{m} \). The sampling sector \( \mathcal{R} \) has radius \( r = 30 \, \text{m} \) and angle \( \theta = 45^\circ \), and the default value for the probability \( \gamma \) is set to \( 0.85 \). We should mention here that the output of the VLM-RRT algorithm is a path of length \( \ell \), i.e., a sequence of \( \ell \) points, which is converted to the continuous reference path \( \mathcal{P} \) to be tracked by fitting a spline curve~\cite{wang2002arc}. The planning horizon is set to \( T = 2.5\ell \), and subsequently, \( T \) evenly spaced points (i.e., with equal arc-length spacing) are sampled from \( \mathcal{P} \) between the starting and goal points. The optimization in Eq.~\eqref{eq:MPC} is solved as a quadratic program (QP) with the Gurobi solver, with the matrices \( Q \) and \( R \) set to \( 0.9 I_{2 \times 2} \) and \( 0.1 I_{2 \times 2} \), respectively.

We have evaluated the performance of our proposed approach by integrating two state-of-the-art VLMs to support autonomous navigation: OpenAI's GPT-4o~\cite{achiam2023gpt} and Meta's Llama 3.2 90B Vision Instruct~\cite{dubey2024llama}. GPT-4o is a multimodal model with approximately 1.8 trillion parameters, accessed via OpenAI's API, whose advanced vision-language processing capabilities were leveraged for environmental interpretation and decision support within our UAV navigation framework. In contrast, Meta's Llama 3.2 90B Vision Instruct, with 90 billion parameters, facilitates multimodal reasoning to provide visual understanding and contextual analysis critical to the VLM-RRT path-planning algorithm.

\begin{table}
\centering
\caption{Performance Comparison with Competing Approaches.}
\resizebox{\columnwidth}{!}{ 
\begin{tabular}{l c c c}
\hline
\textbf{Algorithm} & \textbf{LLM} & \textbf{Avg. Iterations ($N$)} & \textbf{Avg. Path Length (m)} \\
\hline
\addlinespace
A* \cite{Hart1968}       & -         & 514 & 52.73 \\
\addlinespace
\multirow{2}{*}{LLM-A* \cite{meng2024llm}} & GPT-4o     & 410 & 52.73 \\
                        & Llama 3.2V & 405 & 52.80 \\
\hline
\addlinespace
RRT      & -         & 423 & 56.48 \\
RRT* \cite{karaman2011sampling}    & -         & 477 & 53.89 \\
\addlinespace
\multirow{2}{*}{VLM-RRT} & GPT-4o     & 172  & 54.56 \\
                        & Llama 3.2V & 176  & 55.87 \\
\hline
\end{tabular}
}
\label{tab:Baseline_table}
\end{table}

\begin{table*}[t]
\centering
\caption{Performance of VLM-RRT under various models and prompting techniques.}
\resizebox{\textwidth}{!}{ 
\begin{tabular*}{\textwidth}{@{\extracolsep{\fill}} l c c c c c}
\hline
\textbf{Algorithm} & \textbf{VLM} & \textbf{Prompt Technique} & \textbf{Success Rate} & \textbf{Avg. Iterations ($N$)} & \textbf{Avg. Path Length (m)} \\
\hline
\addlinespace
RRT  & - & - & 82\% (41/50) & 343 & 58 \\
RRT* & - & - & 88\% (44/50) & 302 & 45 \\
\midrule
\multirow{3}{*}{VLM-RRT (Ours)} & \multirow{3}{*}{GPT-4o} & Zero-shot & 68\% (34/50) & 93 & 47 \\
 & & Few-shot & 94\% (47/50) & 89 & 46 \\
 & & CoT      & 86\% (43/50) & 94 & 48 \\
\cmidrule{2-6}
 & \multirow{3}{*}{Llama 3.2V} & Zero-shot & 56\% (28/50) & 102 & 48 \\
 & & Few-shot  & 90\% (45/50) & 88 & 49 \\
 & & CoT       & 78\% (39/50) & 95 & 47 \\
\hline
\end{tabular*}
}
\label{tab:VLM-RRT_Table}
\end{table*}

\subsection{Results}

We begin the evaluation by comparing the proposed approach with the closely related work in~\cite{meng2024llm}, where the authors integrated LLMs with the A* path-finding algorithm~\cite{Hart1968}. Additionally, we compare it with the traditional RRT approach~\cite{lavalle1998rapidly} and the RRT* algorithm~\cite{karaman2011sampling}. Table~I presents the average number of iterations required for convergence and the resulting path length for each approach. These results were obtained by averaging 100 random scenarios (i.e., random environment configurations) in a Monte Carlo (MC) simulation.
It is important to note that while the A* and RRT approaches are not directly comparable (in terms of the number of iterations), the results indicate a consensus on the performance improvement achieved when these planning algorithms are integrated with LLMs. In particular, the VLM-RRT approach significantly improves the convergence rate compared to the original RRT algorithm while also enhancing path quality (in terms of path length). Moreover, the VLM-RRT approach achieves path quality comparable to that of the more advanced RRT*, but with fewer iterations. An illustrative example is shown in Fig.~\ref{fig:Baseline_image}. These results were obtained using CoT prompting.

\begin{figure*}
    \centering
    \includegraphics[width=1.0\textwidth]{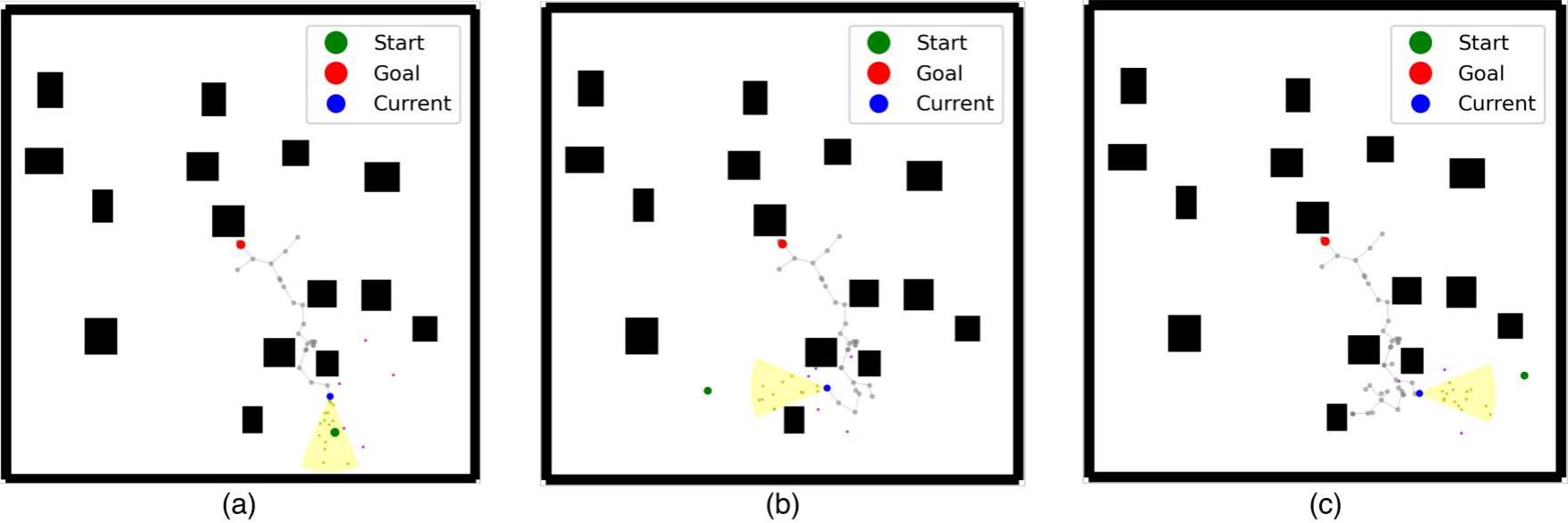}
    \caption{Illustration of the VLM-RRT algorithm navigating toward dynamic goals in a 2D environment. The scenario involves three goal relocations. Red point is the starting position, green point is the goal position which  changes over time, and the blue point is a leaf node. The yellow sector is the VLM-informed sampling region.}
    \label{fig:dynamic_location}
\end{figure*}

In the next experiment, we conduct a more thorough analysis of the proposed VLM-RRT approach in terms of the following metrics:
\begin{enumerate}
    \item \textbf{Success Rate:} Defined as the percentage of experiments in which a collision-free path from the start to the goal region is successfully found within the maximum number of iterations \( N = 500 \).
    \item \textbf{Number of Iterations:} The total number of iterations required to obtain a feasible path.
    \item \textbf{Path Length:} Assessed via the length of the final path from the start to the goal region.
\end{enumerate}
The above metrics are computed on a per-experiment basis and then averaged across 250 MC runs, as shown in Table~II. As shown in the results, RRT and RRT* achieve success rates of 82\% and 88\%, respectively. In contrast, VLM-RRT surpasses the performance of the traditional approaches, particularly under few-shot and CoT prompting, while requiring significantly fewer iterations. Additionally, VLM-RRT achieves higher path quality in terms of path length compared to the traditional RRT, as shown in Table~II. In terms of path length, VLM-RRT achieves performance comparable to the more advanced RRT*, though this is accomplished with fewer iterations. The results also highlight the differences in performance obtained using different prompting techniques, as well as the effectiveness of the two LLMs, indicating an advantage for GPT-4o.

\begin{table}
\centering
\caption{VLM-RRT robustness analysis}
\resizebox{\columnwidth}{!}{ 
\begin{tabular}{c c c}
\hline
\boldmath{$\gamma$} & \textbf{Success Rate (\%)} & \textbf{Avg. Number of Iterations ($N$)} \\
\hline
1.0  & 79  & 86 \\
0.9  & 88  & 92 \\
0.8  & 92 & 105 \\
0.7  & 95  & 102 \\
0.6  & 93  & 128 \\
0.5  & 89  & 142 \\
\hline
\end{tabular}
}
\label{tab:sampling_results}
\end{table}

The next experiment investigates how the parameter \( \gamma \) (i.e., the probability of taking a VLM-informed decision) affects the sensitivity of the algorithm and how this can be fine-tuned to increase the robustness of VLM-RRT. Table~\ref{tab:sampling_results} shows the algorithm's performance (in terms of success rate and number of iterations) for different values of \( \gamma \) ranging from 1 to 0.5, obtained over 100 MC trials. When \( \gamma = 1 \), VLM-RRT always takes a VLM-informed decision at each time step and samples a new point from within the area suggested by the VLM. Although this can lead to faster convergence in many situations, the results show that this strategy decreases robustness (i.e., increases the likelihood of failure to converge). From our experiments, we have observed that this is due to two main reasons: (a) the VLM can make mistakes, which consequently lead the algorithm to make incorrect decisions, and (b) \( \gamma = 1 \) leads to greedy behavior, which in turn causes the algorithm to become stuck in infeasible regions. On the other hand, a lower value of \( \gamma \) causes the algorithm to behave more similarly to the original RRT, resulting in a drop in success rate due to reaching the maximum number of iterations. Therefore, a fine-tuned value of \( \gamma \) optimally balances exploration and exploitation, leading to enhanced robustness and performance, as shown.

In emergency response scenarios, mission parameters often change dynamically as new information becomes available. For instance, the location of survivors may be updated based on new sensor data or witness reports, requiring rapid replanning of UAV trajectories. To evaluate our system's performance in such dynamic scenarios, we conducted a series of experiments with dynamic goal locations. Figure~\ref{fig:dynamic_location} shows one such scenario where the goal region changes location dynamically during the mission.
The VLM demonstrated a consistent ability to identify changes in the location of the goal region during the mission, achieving a detection rate of 92\% across 50 random scenarios in which the goal region's location varied over time. In cases where the VLM failed to immediately recognize the new goal location, it typically required one additional sampling iteration to correct its direction.
We observed that the VLM's performance in recognizing and adapting to new goal locations remained robust even in cluttered environments, though the convergence time increased when obstacles were present between the UAV's position and the new goal location. This increase in convergence time was primarily due to the necessary local path adjustments around obstacles rather than any delay in goal recognition or sampling direction updates.

\section{Conclusion and Future Work}
\label{sec:Conclusion}
In this work, we present Visual-Language Model RRT (VLM-RRT), a hybrid path-planning framework that combines the pattern recognition capabilities of Vision-Language Models (VLMs) with the efficiency of Rapidly-exploring Random Trees (RRT). By utilizing VLMs to extract high-level semantic information from environmental snapshots, our approach directs sampling toward regions with a higher likelihood of containing feasible paths. This targeted bias significantly enhances both sampling efficiency and path quality. Our experimental evaluation demonstrates notable improvements in navigation performance compared to conventional sampling-based methods, highlighting the advantage of integrating VLM-based perception-driven guidance into motion planning. Future work will explore tighter integration between large language models (LLMs) and path-planning algorithms, focusing on how recent advances in reasoning models can further enhance autonomous decision-making and planning capabilities.
\section*{Acknowledgments}
This work is implemented under the Border Management and Visa Policy Instrument (BMVI) and is co-financed by the European Union and the Republic of Cyprus (GA:BMVI/2021-2022/SA/1.2.1/015), and supported by the European Union's Horizon 2020 research and innovation programme under grant agreement No 739551 (KIOS CoE), and through the Cyprus Deputy Ministry of Research, Innovation and Digital Policy of the Republic of Cyprus.

\flushbottom
\balance

\bibliographystyle{IEEEtran}
\bibliography{main}

\end{document}